\documentclass[journal]{IEEEtran}
\usepackage{amsfonts}
\usepackage{amsmath}
\usepackage{amssymb}
\usepackage{pifont}
\usepackage{nicefrac} 
\usepackage{algorithm}
\usepackage{algorithmicx}
\usepackage{algpseudocode}
\usepackage{dsfont}
\usepackage{authblk}
\usepackage{bbm}
\usepackage{graphicx}
\usepackage{epstopdf}
\usepackage{subfigure}
\usepackage{stfloats}
\usepackage{eqnarray}
\usepackage{makecell}
\usepackage{multirow}
\usepackage{enumerate}
\usepackage{booktabs}
\usepackage{cite}
\usepackage{balance}
\usepackage{color}
\usepackage{bm}
\usepackage{caption}
\usepackage{mathtools}
\usepackage{url}
\usepackage[table]{xcolor}

\usepackage{rotating}
\usepackage{multirow}
\usepackage{ragged2e}  % 支持 \justifying
\usepackage{array}
\usepackage{makecell}
\usepackage{textcomp}

\begin{document}
	\title{Fine-Tuning and Deploying Large Language Models Over Edges: Issues and Approaches}
	\author{\mbox{Yanjie Dong},
		\mbox{Haijun Zhang,~\IEEEmembership{Fellow, IEEE}, Chengming Li},
		\mbox{Song Guo,~\IEEEmembership{Fellow, IEEE}},
		\mbox{Victor C. M. Leung,~\IEEEmembership{Life Fellow, IEEE}}, \mbox{and Xiping Hu}
%	(\emph{Corresponding author: Xiping Hu})
	\thanks{Y. Dong, C. Li, V.~C.~M.~Leung, and X. Hu are Shenzhen \mbox{MSU-BIT} University, Shenzhen, China. H. Zhang is with University of Science and Technology Beijing, Beijing, China. S. Guo is with The Hong Kong University of Science and Technology, Hong Kong, China.}}
    \maketitle
    \begin{abstract}
	Since the release of GPT2-1.5B in 2019, the large language models (LLMs) have evolved from specialized deep models to versatile foundation models. 
	While demonstrating remarkable zero-shot ability, the LLMs still require fine-tuning on local datasets and substantial memory  for deployment over the network edges. 
	Traditional first-order fine-tuning techniques require significant GPU memory that exceeds the capacity of mainstream hardware.
	Besides, the LLMs have been expanded beyond text generation to create images, audio, video, and multi-modal content, necessitating careful investigation of efficient deployment strategies for large-scale foundation models. 
	In response to these challenges, model fine-tuning and model-compression techniques have been developed to support the sustainable growth of LLMs by reducing both operational and capital expenditures.
	In this work, we provide a comprehensive overview of prevalent memory-efficient fine-tuning methods for deployment at the network edge. 
	We also review state-of-the-art literature on model compression, offering insights into the deployment of LLMs at network edges.
    \end{abstract}

%{\color{blue}
%	\section*{Nomenclature}
%	\begin{tabbing}
%		\textbf{Acronyms} \quad\:\= \textbf{Definitions} \\
%		GPT2 \> Generative Pretrained Transformer 2\\
%		LLM \> Large Language Model \\
%		BP \> Backward Propagation \\
%		GPU \> graphics process unit \\
%		LLAMA \> Large Language Model Meta AI \\
%		GB \> Gigabyte \\
%		FP16 \> 16-bit Floating Point \\
%		FP32 \> 32-bit Floating Point \\
%		PEFT \> Parameter-Efficient Fine-Tuning \\
%		MEFT \> Memory-Efficient Fine-Tuning \\
%		DLNoE \> Distributed Learning Networks Over Edge \\
%		LoRA \> Low-Rank Adaption \\
%		FedPepTAO \> Federated Parameter-Efficient Prompt \\
%					\>\quad\quad Tuning With Adaptive Optimization \\
%		FedPETuning \> Federated Parameter-Efficient Tuning \\
%		ZO \> Zeroth-Order Optimizer \\
%		ZO-SGD \> Zeroth-Order Stochastic Gradient Descent \\
%		MeZO \> Memory-Efficient Zeroth-Order Optimizer \\
%		FedKSeed \> Federated Fine-Tuning With $K$ Seeds \\
%		FwdLLM \> Federated Fine-Tuning LLM With \\
%				\>\quad\quad  Forward Passes\\
%		PS \> Perturbation Size \\
%		KL \> Kullback-Leibler \\
%		GKD \> Generalized Knowledge Distillation \\
%		GJS \> Generalized Jensen-Shannon \\
%		FLAN \> Fine-Tuned Language Net \\
%		OPTQ \> Open Pre-Trained Transformer Quantization 
%\end{tabbing}}

{\color{blue}
\section{Introduction}
Since the OpenAI researchers released the GPT2--1.5B\footnote{GPT2--1.5B denotes a variant of the GPT2 model with 1.5 billion weights. Hereinafter, we refer to each large language model by the name and the corresponding model size.} to the world in 2019, the language models have shifted from specialized and fragmented deep models to versatile and one-size-fits-all large foundation models.
Earlier investigations have suggested that the large language models (LLMs) exhibit high-level generalization that allows for transferring acquired knowledge to unseen tasks (a.k.a., zero-shot ability), such as, sentiment analysis, machine translation, and question answering \cite{Chen2021}.
Amid the trend toward scaling to hundreds of billions of weights, the LLMs have demonstrated a remarkable ability to generate coherent and contextually relevant text that has sparked widespread interests and driven significant advancements in the field of generative artificial intelligence. 
Alongside the proliferation of LLMs, large foundation models have also been extended to generate images, audio, video, and multimodal content.  
Despite their impressive zero-shot capabilities, domain adaptation via \emph{fine-tuning} is typically required to align model behavior with the specific datasets of downstream tasks \cite{Wu2024}.
Besides, the sheer scale of model weights imposes prohibitive memory, latency, and energy costs for the edge devices to deploy versatile LLMs. 
Therefore, \emph{model-compression} techniques have become indispensable companions to fine-tuning and offer complementary ways to reduce computational and storage overhead without sacrificing the performance of LLMs.

\paragraph{\bfseries Memory Bottlenecks During Fine-Tuning}Fine-tuning LLMs is conventionally performed with the first-order optimizers (e.g., Adam and AdaGrad), which compute gradients via the backward propagation (BP) operations.
However, the BP operations consume several times more graphics processing unit (GPU) memory compared to the inference operations \cite{Malladi2023}.
For example, the \mbox{LLAMA--7B} consumes roughly 14GB of memory per inference with weights in 16-bit floating point (FP16) format. 
Fine-tuning the \mbox{LLAMA--7B} via the Adam optimizer requires approximately 28GB of memory to store model weights and gradients in FP16 format.
When performing BP operations, the Adam optimizer requires extra memory to store momentums (28GB of memory) and variances (28GB of memory) in 32-bit floating point (FP32) format.
When activation checkpoints are included, the memory requirement for fine-tuning \mbox{LLAMA--7B} via the Adam optimizer exceeds 84GB of memory that surpasses the memory capacity of mainstream GPUs (e.g., NVIDIA A800 and RTX 4090).
The huge memory requirement and the high capital expenditure\footnote{As of 2024, the prices of the Nvidia A800 and RTX 4090 are approximately 22,400 USD and 1,599 USD, respectively.} have catalyzed research into the novel model fine-tuning techniques, which fall broadly into two categories: (i) \textbf{parameter-efficient fine-tuning (PEFT)} that learns a small subset of weights while freezing the remaining backbone, and  (ii) \textbf{memory-efficient fine-tuning (MEFT)} that replaces the BP operations with the memory-light alternatives.

\paragraph{\bfseries Synergy Between Fine-Tuning and Model Compression}Whereas fine-tuning adapts an LLM to task-specific data, the model compression transforms the adapted LLM into a resource-efficient artifact that is suitable for deployment under stringent latency, power, and memory constraints. 
These two processes are deeply interrelated. 
PEFT and MEFT techniques alleviate memory and computational bottlenecks during fine-tuning of LLMs; consequently, the LLMs can be adapted on resource-constrained edge devices.
Meanwhile, the model-compression techniques (e.g., pruning, knowledge distillation, and quantization) can be applied to reduce the scale and maintain the versatility of LLMs.
Besides, model-compression techniques can be applied \emph{during} and \emph{after} fine-tuning to further reduce model size and inference cost. 
Such complementary synergy not only facilitates cost-effective and scalable edge deployment of LLMs but also contributes to the broader goal of sustainable AI by significantly lowering energy consumption and carbon footprint in the development and deployment of LLMs. 
These methodologies collectively constitute a promising path toward democratizing access to powerful LLMs across diverse application scenarios and hardware platforms.

\paragraph{\bfseries Contributions} Based on the above insights, we review recent research advances on fine-tuning and model-compression techniques over edges.
Note that the abundance of private data on edge devices has driven the integration of LLM fine-tuning with distributed learning networks over edge (DLNoE) though such integration poses significant computational challenges.
Therefore, we begin by reviewing the PEFT and MEFT techniques, and outline the most promising solutions. 
Then, we review the state of the arts in model-compression for DLNoE. 
Our major contributions are twofold: \textbf{(i).~Fine-tuning on edge devices.} 
For obtaining task-specific LLMs over edges, we synthesize exemplary PEFT and MEFT works that substantially reduce memory footprints due to the memory bottleneck of fine-tuning in DLNoE. 
We further present a numerical evaluation demonstrating the efficacy of federated MEFT under typical DLNoE settings.
\textbf{(ii).~Model-compression taxonomy.} 
For obtaining versatile LLMs over edges, we categorize existing literature into three paradigms (i.e., \emph{compression-and-train}, \emph{compression-then-train}, and \emph{one-shot compression}) and systematically analyze representative algorithms of the three categories.

%The remaining work is organized as follows. 
%Section II and III respectively discuss the issues and approaches of federated PEFTs and MEFTs in the DLNoE. 
%Section IV discusses the state of-the-art model-compression techniques for edge devices. 
%Since the DLNoE for LLMs are at the early stage of investigation, Section V concludes our work and provides several future directions.
}

\section{Parameter-Efficient Fine-Tuning in DLNoE}
\begin{figure}[htb]
	\centering
	\includegraphics[width=0.9\linewidth]{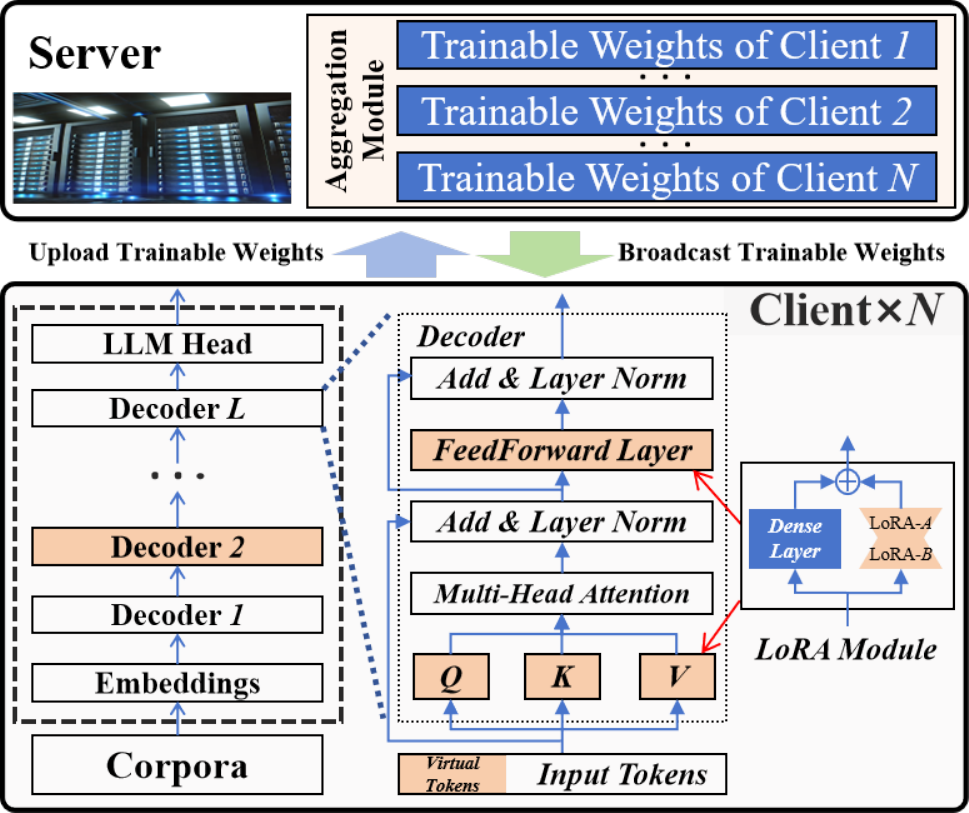}
	\caption{\color{blue} Federated PEFT techniques in the $N$-client DLNoE where LoRA matrices \cite{Zhang2024a}, inserted virtual tokens \cite{Che2023}, and a decoder layer \cite{Sun2025} are fine-tuned.}\label{lbl:peft_scenario}
\end{figure}

Since full-parameter fine-tuning with first-order optimizers requires several-fold more GPU memory than inference, the PEFT methods have emerged to shrink the trainable weights and make fine-tuning LLMs feasible on resource-constrained edge devices.
In the DLNoE, the PEFT techniques can also reduce the communication overhead during fine-tuning due to the reduced size of trainable weights.
Based on ways to obtain the small trainable modules, the PEFT techniques can be classified as reparameterization-based PEFT, addition-based PEFT, and selective PEFT \cite{Ding2023}. 
As shown in Fig. \ref{lbl:peft_scenario}, the reparameterization-based PEFT can replace the dense layers of the LLMs by two trainable low-rank adaption (LoRA) matrices that have much smaller number  of weights than the original dense layers while the addition-based PEFT prepends the trainable modules to transformer blocks of the LLMs.
Instead of introducing extra trainable weights that can increase the model complexity, selective PEFT can update a small set of weights of the LLMs and freeze the remaining backbone.
Hereinafter of this section, we review recent advances in the three PEFT categories suited to DLNoE deployment.

{\color{blue}
\paragraph{\bfseries Reparameterization-Based PEFT in DLNoE}Motivated by the premise that fine-tuning for most downstream tasks is intrinsically low-rank \cite[and references therein]{Ding2023}, the LoRA modules can be used to reparameterize the dense layers (e.g., query, key, value, and feedforward layers) of the LLMs as illustrated in Fig. \ref{lbl:peft_scenario}.
In this vein, the federated instruction-tuning (FedIT) technique assigns each client an LLM and reparameterizes all its dense layers using LoRA modules \cite{Zhang2024a}. 
Note that the trainable LoRA modules in \cite{Zhang2024a} account for only 0.26\% of the total parameters, which renders the LoRA modules particularly suitable for training on resource-constrained edge devices.
During the fine-tuning process, each client utilizes the Adam optimizer to perform local updates and transmits only the trainable weights of the LoRA modules to the central server for aggregation. 
Since each LoRA module consists of two matrices with a  much lower rank than that of the corresponding dense layer, the communication overhead in FedIT is significantly reduced relative to schemes that exchange the full model parameters.
A further advantage of the FedIT technique lies in its ability to accommodate heterogeneous instruction data from geographically distributed clients, which may possess varying domain-specific expertise. 
The numerical experiments in \cite{Zhang2024a} adopt the Databricks-dolly-15k dataset, which contains diverse instruction-following samples across multiple categories.
By partitioning the Databricks-dolly-15k across clients to simulate data heterogeneity, FedIT is shown to improve the generalization of fine-tuned LLMs.

\paragraph{\bfseries Addition-Based PEFT in DLNoE}Soft-prompt PEFT techniques can prepend trainable virtual tokens (a.k.a., soft prompts or continuous prompts) to the input of each transformer layer of LLMs.
Therefore, soft-prompt PEFT can be integrated with the DLNoE to leverage the local private data of individual clients.
However, directly exchanging the trainable weights between clients and the server induces substantial communication overhead.
To alleviate the communication bottleneck, the federated parameter-efficient prompt tuning with adaptive optimization (FedPepTAO) technique \cite{Che2023} introduces an intelligent selection mechanism that dynamically selects exchanged trainable prefix weights between the server and clients.
The selection procedure is based on the hidden-state correlation matrices and Hessian matrices of the prompt weights of all clients.
Upon completion of a fixed number of local training rounds, the server randomly selects a subset of clients and transmits to them the prefix weights of the layers that are identified by the intelligent selection mechanism.
Then, the selected clients update the prefix weights and upload the prefix weights of the selected layers to the server.
Besides, the adaptive first-order optimizer is developed to handle client drift due to data heterogeneity.
More specifically, the adaptive first-order optimizer allows the server and clients to use respectively momentum-based and adaptive optimization techniques.
Such a combination helps in stabilizing the training process and achieving superior performance over the integration of soft-prompt tuning with federated learning as highlighted  in \cite{Che2023}.

\paragraph{\bfseries Selective PEFT in DLNoE}
Without adding extra weights to the target LLMs, the federated selective fine-tuning (FedSelFT) technique was proposed in \cite{Sun2025} by allowing the clients to fine-tune a subset of layers. 
In the FedSelFT technique, the server and all clients maintain identical copies of the target LLMs.
After several rounds of local training, each client uploads the weight drift of pre-selected layers to the server.
The server aggregates the weight drifts across all layers from clients and updates the global model weights based on the aggregated weight drift.
Note that the server can decide the number of fine-tuning layers based on the local resources (e.g., computational power, communication bandwidth, and GPU memory) per client and dynamically choose the subset of layers based on the aggregate gradient variance that is penalized by the selection discrepancy in FedSelFT.
Therefore, the dual heterogeneity--namely, device and data heterogeneity--in DLNoE can be more effectively addressed compared to traditional static layer selection mechanisms.}

\paragraph{\bfseries A Unified Federated PEFT Framework in DLNoE}
A unified framework (namely, FedPETuning) is designed to integrate multiple PEFT techniques into the DLNoE \cite{Zhang2023}. 
In the FedPETuning framework, the clients leverage the three categories of PEFT techniques to fine-tune only a small subset of LLM weights and exchange with the server only the updated LLM weights.
Therefore, the computational and communication overhead can be significantly reduced at the edge devices. 
Different DLNoE scenarios are used to verify the effectiveness of the FedPETuning framework, i.e., the cross-silo and large-scale cross-device scenarios. 
The cross-silo DLNoE usually has smaller number of clients ($\leqslant$100 clients) and more data samples per client than the large-scale cross-device.
By leveraging the Robert-Base-125M model for fine-tuning, the FedPETuning framework can significantly reduce the communication overhead and attain a comparable performance (at most 7\% performance loss) over the federated full-parameter fine-tuning in both cross-silo and large-scale cross-device DLNoEs.
Since the FedPETuning framework allows a small fraction of weights to be exchanged between the server and clients, the privacy leakage can also be reduced by increasing the batch size of fine-tuning samples \cite{Zhang2023}.

{\color{blue}
The PEFT techniques can significantly reduce the scale of trainable weights such that fine-tuning tasks can be executed directly on resource-constrained edge devices in the DLNoE.
By using respectively reparameterization-based FedIT \cite{Zhang2024a, Zhang2023}, addition-based FedPepTAO \cite{Che2023, Zhang2023}, and selective trainable \cite{Sun2025, Zhang2023} modules, the FedIT \cite{Zhang2024a}, FedPepTAO \cite{Che2023}, FedSelFT \cite{Sun2025} selectively exchange subsets of trainable weights.
However, the communication overhead remains inevitably high due to the frequent exchanges of weights and gradients. 
Moreover, the federated PEFT techniques still incur performance loss compared to full-parameter fine-tuning \cite{Zhang2023}. 
Additionally, the federated PEFT techniques typically rely on BP-based first- and second-order optimizers, yet the BP operations are seldom feasible on system-on-chip edge devices, such as Qualcomm Snapdragon and Huawei Kirin.
When edge devices are unable to support efficient BP operations, an alternative research direction explores BP-free zeroth-order (ZO) optimizers and will be detailed in the next section. 
}

\section{Memory-Efficient Fine-Tuning in DLNoE}
As an exemplary case, the zeroth-order stochastic gradient descent (ZO-SGD) achieves gradient estimation by performing two loss evaluations and a single stochastic perturbation, which eliminates the need for BP operations.
The scalar gradient and the stochastic perturbation are then combined as the zeroth-order gradient (ZOG) for model update via vanilla SGD. 
{\color{blue} However, ZO-SGD still incurs a memory footprint approximately twice that of inference since each iteration must store the model weights and the applied perturbation.}
Therefore, the ZO-SGD optimizer is inefficient for fine-tuning LLMs.
To address such limitation, Malladi \emph{et al.} proposed the memory-efficient zeroth-order (MeZO) optimizer \cite{Malladi2023}. 
By resetting the random seed, MeZO can regenerate stochastic perturbations and leverage the in-place weight updates without incurring additional GPU memory.
Therefore, MeZO can fine-tune substantially larger models compared to BP-based optimizers. 
{\color{blue} For example, MeZO can fine-tune a 30B model on a single GPU with 80GB of memory, whereas \mbox{BP-based} optimizers are limited to around 2.7B.}
Theoretically, the convergence rate of MeZO is tied to the local effective rank of the Hessian that is typically much smaller than the number of weights. 
MeZO also supports non-differentiable objectives and is suitable for fine-tuning tasks that involve human feedback and safety constraints.

\begin{figure}[htb]
\centering
\includegraphics[width= \linewidth]{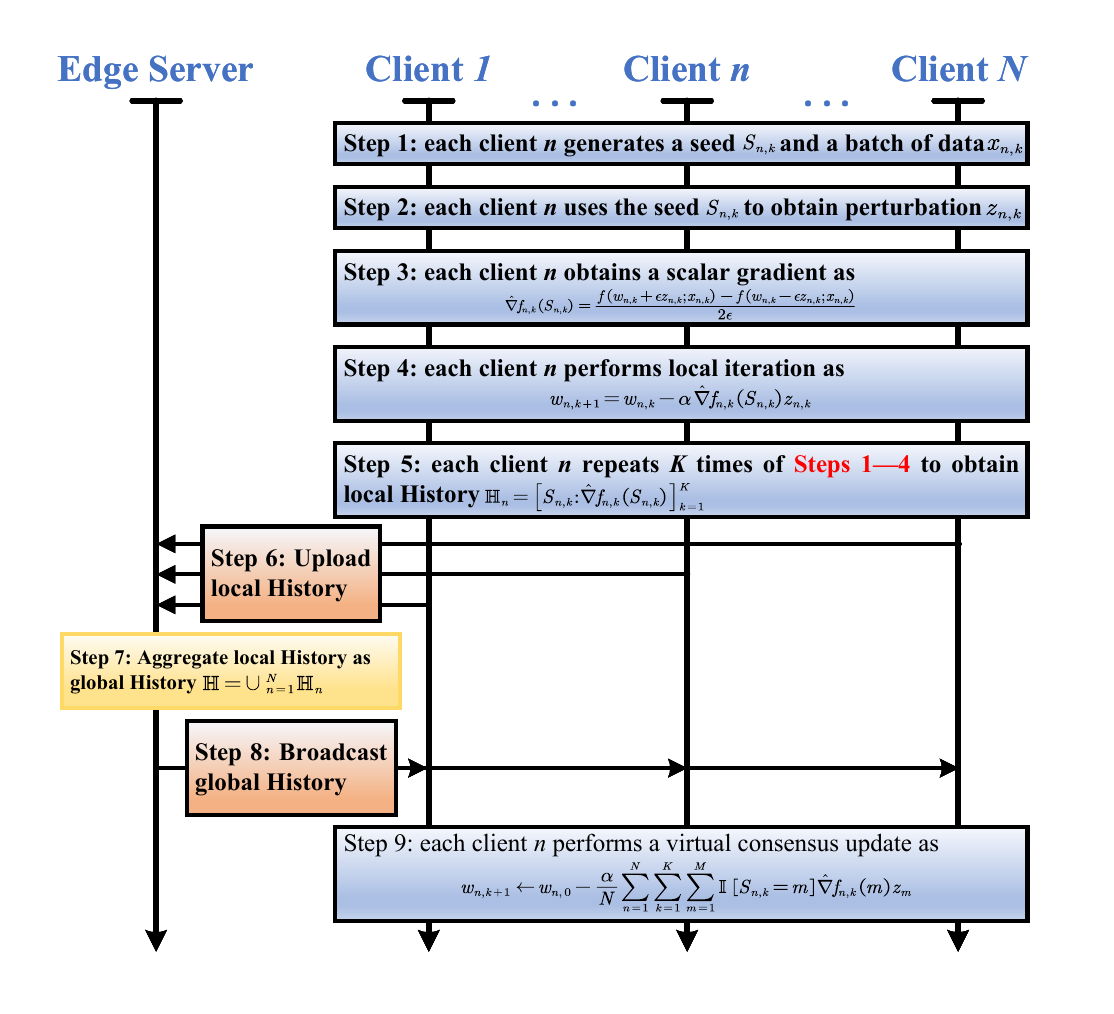}
\caption{\color{blue}Federated MEFT procedures in the $N$-client DLNoE that uses either infinite \cite{Zelikman2023, Xu2024b} or finite \cite{Qin2023} numbers of local seeds.}\label{lbl:fedzo}
\vspace{-0.6 cm}
\end{figure}

{\color{blue}
\paragraph{\bfseries Enabling Communication- and Memory-Efficient DLNoE}Given the prohibitive communication overhead of federated PEFTs and the aforementioned merits, Eric Zelikman \emph{et al.} extended the MeZO to the federated setting as just-one-byte (JOB) that can perform \emph{full-parameter} fine-tuning in the DLNoE and remarkably reduce the communication overhead compared to federated PEFTs \cite{Zelikman2023}.
To enhance the scalability of client addition and removal, each new client is initialized with the current iteration index, and all active clients are notified of its arrival in order to perform the subsequent synchronized fine-tuning procedures.
In the JOB technique, each client performs several rounds of local ZO-SGD via the random seed that is constructed from the client identifier time, the current iteration index, and the sample batch index.
Then, the server aggregates a global seed-to-gradient (SG) dictionary that contains random seed and scalar gradient pair of all clients and broadcasts such dictionary to all clients. 
By leveraging the global SG dictionary, each client can reproduce the stochastic perturbations that correspond to the scalar gradients using the associated seeds.
Finally, all clients obtain the up-to-date model weights via the reproduced stochastic perturbations and scalar gradients.

Although JOB can reduce communication overhead by exchanging the SG dictionary, the potentially infinite number of random seeds can induce growing computation costs over communication rounds.
Therefore, Zhen Qin \emph{et al.} proposed the FedKSeed that inherits the strengths of JOB and alleviates its growing computational expenditure by restricting the random seeds to a finite set \cite{Qin2023}.
As shown in Fig. \ref{lbl:fedzo}, FedKSeed allows for the server to broadcast the global SG dictionary to clients.
Based on the received global SG dictionary, all clients compute the latest global model, perform local training, and upload the 
local SG dictionaries to the server.
By restricting perturbation generation to a finite set of random seeds, the FedKSeed expedites the convergence of fine-tuning LLMs with low local computational expenditure.
Theoretical analysis and numerical experiments further indicate that, when the finite seed set is sufficiently large, its effect on fine-tuned LLMs is negligible \cite{Qin2023}.
Due to the extraordinary capability for fine-tuning LLMs  on resource-constrained edge devices, the FedKSeed is envisioned as a promising communication- and memory-efficient fine-tuning technique in DLNoE.

\paragraph{\bfseries Fusing PEFT with MEFT in DLNoE}Motivated by the highly-efficient inference of neural processing units on edge devices, Mengwei Xu \emph{et al.} proposed the FwdLLM that integrates MEFT with PEFT techniques \cite{Xu2024b}.
Since single perturbation induces high variance of the ZOG, the FwdLLM employs multiple perturbations. 
However, increasing the number of perturbations improves ZOG estimation accuracy at the cost of higher local inference overhead.
To balance computation-accuracy tradeoff, FwdLLM confines the total number of perturbations aggregated from all clients per iteration.
Per each iteration, the server dynamically controls the number of perturbations by setting a variance threshold to scalar gradients in the global SG dictionary while the clients are responsible for uploading the random seed and scalar gradient pairs to the server. 
When the variance of scalar gradients drops below the predefined threshold, the server updates the trainable weights in the LLM and notifies the clients with the up-to-date LLMs. 
In FwdLLM, the size of global SG dictionary is primarily increased by involving more devices for parallel computation and faster convergence. 
Once the device limit is reached, FwdLLM further enlarges the global SG dictionary by increasing the number of local perturbations per client.
Additionally, leveraging the smooth variation of gradient directions, an intelligent perturbation mechanism is implemented by selecting the perturbation directions with the top cosine similarities to the previous ZOG to construct the ZOG of current iteration \cite{Xu2024b}.
Based on \emph{dynamic perturbation number} and \emph{intelligent perturbation mechanism}, the FwdLLM can effectively integrate the BP-free optimizer with the PEFT techniques such that the efficient and scalable fine-tuning process for LLMs can be realized on  edge devices.

Compared with the federated PEFTs, the federated MEFTs (e.g., JOB \cite{Zelikman2023}, FedKSeed \cite{Qin2023}, FwdLLM \cite{Xu2024b}) do not need to cache the activations such that the full-parameter fine-tuning of LLMs can be performed at edge devices.
The communication overhead between the server and clients can also be drastically reduced by  exchanging only the SG dictionary with potentially infinite size \cite{Zelikman2023, Xu2024b} or finite size \cite{Qin2023}. 
Moreover, restricting the search directions by limiting random seeds \cite{Qin2023} or employing high-similarity perturbations \cite{Xu2024b} can accelerate the convergence of federated MEFT techniques.
	%However, the scale of LLMs on a single GPU remains limited by its memory capacity.
	%To increase the scale of LLMs on a single GPU, model-compression techniques will be discussed in the next section.
}

{\color{blue}
\begin{table}[hbt]\scriptsize	
	\centering
	\caption{Comparisons of representative fine-tuning techniques for the OPT--6.7B model in FP16 format, evaluated with 128 input tokens per sample and 32 samples per batch.}\label{tab:comparison}%
	\begin{tabular}{|l !{\vrule width 1.5 pt} >{\centering}m{4.9 em}|>{\centering}m{5 em}|>{\centering}m{4.45 em}|c|}
		\Xhline{1.5 pt}
		\rowcolor{cyan!20} \bfseries Techniques &  
		\bfseries Running Memory & 
		\bfseries Commun. Overhead &
		\bfseries Optimizers & 
		\bfseries Scalability 
		\\\Xhline{1.5 pt}
		FedIT \cite{Zhang2024a} &  24--28GB \mbox{\!LoRA-R=8}  &  LoRA weights ($\approx$8--32M)  & First-order  &  Medium \\\hline
		FedPepTAO \cite{Che2023} &  \mbox{24--27.4GB} Prefix Len.=16-64  &  Selected Prefixes ($\approx$2–8M) & First-order  & Medium \\\hline
		FedSelFT \cite{Sun2025} & \mbox{26--29.7GB} \# Layer=1  & Selected Layers ($\approx$200M)  & First-order   & Low  \\\Xhline{1.5 pt}
		JOB \cite{Zelikman2023} & 15--16GB  & (Infinite) \mbox{SG Dict.}  &  MeZO  &  High   \\\hline
		FedKSeed \cite{Qin2023} & 15--16GB  & Finite \mbox{SG Dict.}      &  MeZO  &  High	 \\\Xhline{1.5 pt}
		FwdLLM \cite{Xu2024b}   & 15--16GB  & (Infinite) \mbox{SG Dict.}  &  MP-MeZO  &  High   \\\Xhline{1.5 pt}
	\end{tabular}%
	\vspace{-1.5 mm}
\begin{flushleft}
\scriptsize
$^1$\textit{Commun. Overhead:} communication overhead, \textit{SG Dict.:} SG dictionary 
\end{flushleft}
\vspace{-0.4 cm}
\end{table}%

\paragraph{\bfseries Comparative Analysis}We compare the federated PEFT and MEFT techniques in DNLoE with respect to GPU memory consumption, communication overhead, adopted optimizers, and scalability, as summarized in Table \ref{tab:comparison}. 
The federated PEFTs (i.e., FedIT, FedPepTAO, and FedPETuning) require caching of forward activations during training; consequently, their GPU memory usage is nearly double that of federated MEFTs (i.e., JOB, FedKSeed, and FwdLLM).
Since federated MEFTs exchange only the SG dictionary per each iteration among the clients, their communication overhead (i.e., JOB, FedKSeed, and FwdLLM) is negligible compared to that of federated PEFTs.
As DLNoE scales, the federated PEFTs and MEFTs support dynamic client participation without global reinitialization. 
However, federated MEFTs achieve better scalability since only a global SG dictionary is exchanged, which greatly reduces communication overhead of the server, while PEFT requires frequent exchange of trainable modules, limiting efficiency in large-scale deployments.

\begin{figure*}[ht]
	\vspace{-0.2 cm}
	\centering
	\subfigure{\includegraphics[width=0.26\linewidth]{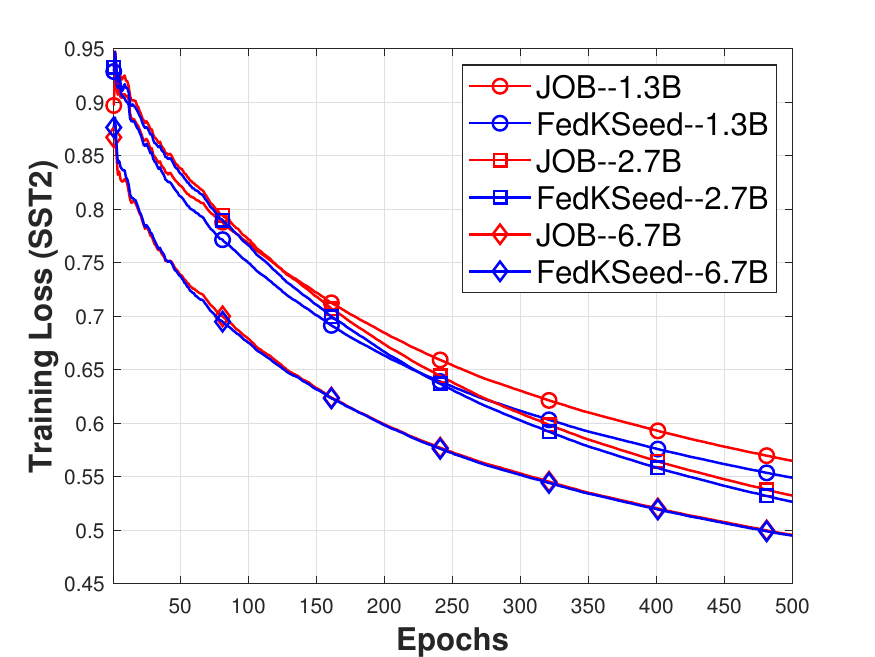}\label{fig:sst}}\hspace{-0.5 cm}
	\subfigure{\includegraphics[width=0.26\linewidth]{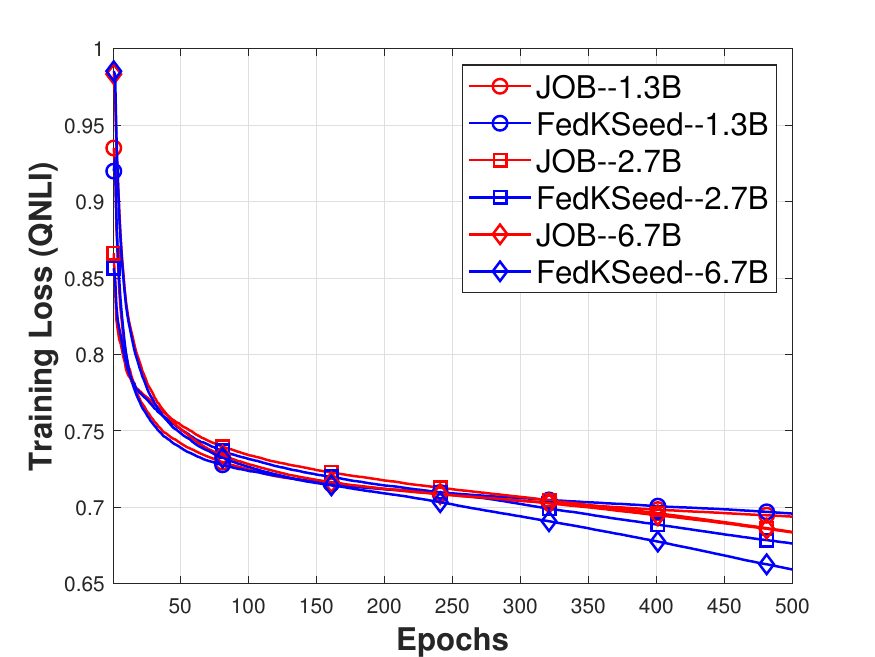}\label{fig:qnli}}\hspace{-0.5 cm}
	\subfigure{\includegraphics[width=0.26\linewidth]{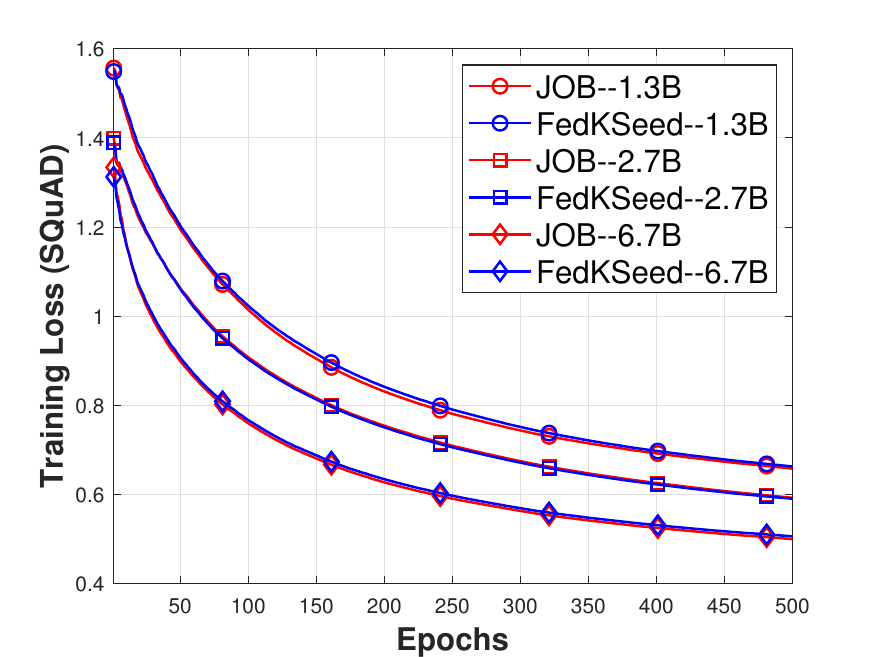}\label{fig:squad}}\hspace{-0.5 cm}
	\subfigure{\includegraphics[width=0.26\linewidth]{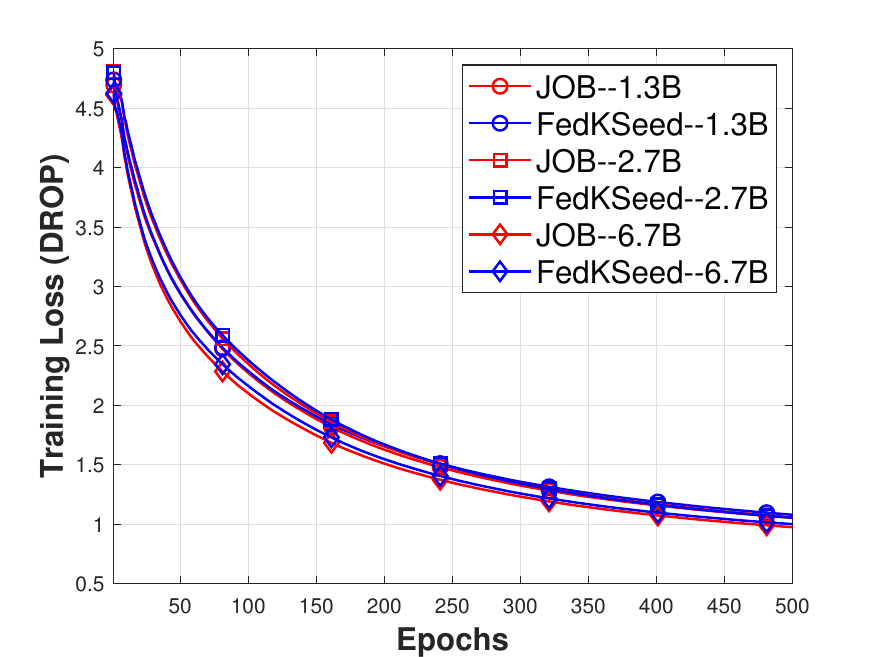}\label{fig:drop}}
	\vspace{-0.3 cm}
	\caption{Running average training loss for JOB and FedKSeed over SST2, QNLI, SQuAD, and DROP datasets.}\label{fig:fedmeft}
	\vspace{-0.0 cm}
\end{figure*}

\begin{table*}[ht]
	\centering
	\begin{tabular}{|l !{\vrule width 1.5 pt} r|r|r|r !{\vrule width 1.5 pt} r|r|r|r !{\vrule width 1.5 pt} r|r|r|r|}
		\Xhline{1.5 pt} 	
		\rowcolor{cyan!20} \bfseries Models 
		& \multicolumn{4}{c !{\vrule width 1.5 pt}}{\bfseries OPT--1.3B}
		& \multicolumn{4}{c!{\vrule width 1.5 pt}}{\bfseries OPT--2.7B} 
		& \multicolumn{4}{c|}{\bfseries OPT--6.7B} \\
		\Xhline{1.5 pt}
		\rowcolor{yellow!20} Tasks 
		& \multicolumn{1}{c|}{SST2} & \multicolumn{1}{c|}{QNLI} & \multicolumn{1}{c|}{SQuAD} & \multicolumn{1}{c !{\vrule width 1.5 pt}}{DROP} 
		& \multicolumn{1}{c|}{SST2} & \multicolumn{1}{c|}{QNLI} & \multicolumn{1}{c|}{SQuAD} & \multicolumn{1}{c !{\vrule width 1.5 pt}}{DROP} 
		& \multicolumn{1}{c|}{SST2} & \multicolumn{1}{c|}{QNLI} & \multicolumn{1}{c|}{SQuAD} & \multicolumn{1}{c !{\vrule width 1.5 pt}}{DROP}  \\\Xhline{1.5 pt}
		Zero-Shot  &   0.5356    &   0.5270    &   0.3040    &  0.1066     &   0.5631    &  0.5280  &  0.3920 & 0.1091     &   0.6124    &   0.5260    &   0.3330    &  0.1173 \\\hline
		JOB 	   &   0.8050    &   0.6310    &   0.5440    &  0.7392     &   0.8578    &  0.7050  &  0.5080 & \bfseries  0.7563     & \bfseries  0.8899    & \bfseries  0.7330    &   0.5980    &  0.6917 \\\hline
		FedKSeed   &  \bfseries 0.8062    & \bfseries  0.6430    &  \bfseries 0.5930    &  \bfseries 0.7396     & \bfseries  0.8681    &  \bfseries 0.7400  & \bfseries  0.5090 & 0.7157     &   0.8842    &   0.7220    &  \bfseries 0.6360  & \bfseries  0.7534 \\\hline
		FedKSeed-LoRA &  0.5367  &   0.5180    &   0.3180    &  0.1120     &   0.5642    &  0.5180  &  0.4130 & 
		0.1160     &   0.6135    &   0.5230    &   0.3960    &  0.1220 \\\Xhline{1.5 pt}
	\end{tabular}%
\vspace{-1.5 mm}
\begin{flushleft}
	\scriptsize
	$^1$Task-wise performance metrics: SST2 -- Accuracy, QNLI -- Accuracy, SQuAD -- Exact Match, DROP -- F1 Score.
\end{flushleft}
\vspace{-0.3 cm}
\caption{Performance of the typical federated MEFT techniques.}\label{tab:fedmeft}%
\vspace{-0.2 cm}
\hrulefill
\vspace{-0.5 cm}
\end{table*}%

\paragraph{\bfseries Case Study of Federated MEFT in DLNoE} We consider a DLNoE with $800$ clients with each sampled at a probability of 5\textperthousand.
We test the performance of federated MEFTs on four practical downstream tasks: SST2, QNLI, SQuAD, and DROP.
All LLMs are loaded in FP16 format. 
The number of clients is $800$ with sampling rate as 5\textperthousand.
Each client respectively selects $128$, $220$, $264$, and $132$ samples for  SST2, QNLI, SQuAD, and DROP datasets.
The learning rate is set to $1 \times 10^{-7}$ for SST-2, SQuAD, and DROP, and $2 \times 10^{-7}$ for QNLI.
The batch sizes are respectively set as $32$, $10$, $12$, and $6$ for SST2, QNLI, SQuAD, and DROP. 
The number of random seeds is set as $20,480$. 
In terms of task complexity, the relative difficulty of the datasets is ordered as: SST $\le$ QNLI $\le$ SQuAD $\le$ DROP.

Figure \ref{fig:fedmeft} shows that the running averages of the JOB and FedKSeed reduce as the number of epochs for SST2, QNLI, SQuAD, and DROP over the OPT-1.3B, OPT-2.7B, and OPT-6.7B.
We observe that across datasets ranging from the simplest SST2 to the most challenging DROP, FedKSeed consistently converges when the number of random seeds is set to $20,480$. 
Compared to JOB, FedKSeed demonstrates faster convergence on the SST2 and QNLI datasets and achieves a comparable convergence rate on the SQuAD and DROP datasets.
Such observations confirm that constraining the number of random seeds (i.e., search directions) can accelerate convergence. 
Moreover, for more complex tasks (e.g., SQuAD and DROP), increasing the number of search directions further enhances convergence.

Table \ref{tab:fedmeft} shows that FedKSeed consistently outperforms JOB and Zero-Shot and achieves the highest performance on most tasks, especially for complex datasets like SQuAD and DROP, as model size increases. 
FedKSeed-LoRA shows limited gains and performs similarly to Zero-Shot and significantly below FedKSeed.
Such observations indicate that LoRA adaptation may be insufficient when the searching directions lack sufficient quality or informativeness.
Notably, FedKSeed demonstrates robustness and scalability across model sizes, particularly with OPT-6.7B, where it attains the best results on three of four tasks. This confirms FedKSeed’s effectiveness in both classification and reasoning tasks.
}

%We perform a case study to evaluate the impacts of number of seeds on the model performance when using MEF2T in DLNoE. 
%We consider a DLNoE with 200 clients with sampling probability of each client as 5\%. 
%We fine tune the LLAMA-1B model over the Databricks-dolly-15k dataset.
%The number of samples per client follows the Dirichlet distribution with parameter $\alpha$ specified during experiments. 
%Each client uses one data sample per iteration and performs 200 iterations per round. 
%The numerical results are illustrated in \ref{lbl:casestudy}. 
%We observe from Fig. \ref{fig:converge} that the testing loss converges after about 120 rounds with the Dirichlet hyper-parameter $\alpha = 0.5, 1, 1.5, 2, 2.5$ and the number of seeds as $700$. 
%Then, we use the fine-tuned LLAMA-1B model to test the Rouge-L F1 score to illustrate the impacts of number of seeds in Fig. \ref{fig:rouge}. 
%From Fig. \ref{fig:rouge}, we can confirm that the number of seeds has little to no effect on the Rouge-L F1 scores. 
%This observation motivates us to use small amount of seeds to expedite the fine-tuning process. 

\section{Compressing LLMs for Edge Devices}
While achieving remarkable performance on downstream tasks, the versatility as general-purpose task solvers is compromized by the fine-tuned LLMs. 
Besides, deploying LLMs on a single GPU requires significant memory resources.
For example, GPT3--175B in FP16 format requires about 350 GB of GPU memory that exceeds the capacity of mainstream consumer GPUs (e.g., Nvidia A800 and RTX 4090). 
Thus, multiple GPUs are required even for inference, which is impractical for edge devices with limited size and resources.
For deploying advanced LLMs as versatile task solvers on edge devices, model compression is essential for reducing memory footprint. 
Although traditional compression techniques (e.g., knowledge distillation, pruning, and quantization) can shrink model size, they usually require costly retraining to restore performance. 
However, retraining GPT-scale models is especially challenging due to substantial computational demands and the limited availability of high-quality, proprietary datasets. 
{\color{blue}
Therefore, efficient compression techniques are needed to \emph{reduce the required data volume} and \emph{shorten the retraining duration} while preserving zero-shot and generalization capabilities of original LLMs.}

\begin{figure*}[ht]
	\vspace{-0.2 cm}
	\centering
	\subfigure[Compress-and-Train]{\includegraphics[width=5.5 cm]{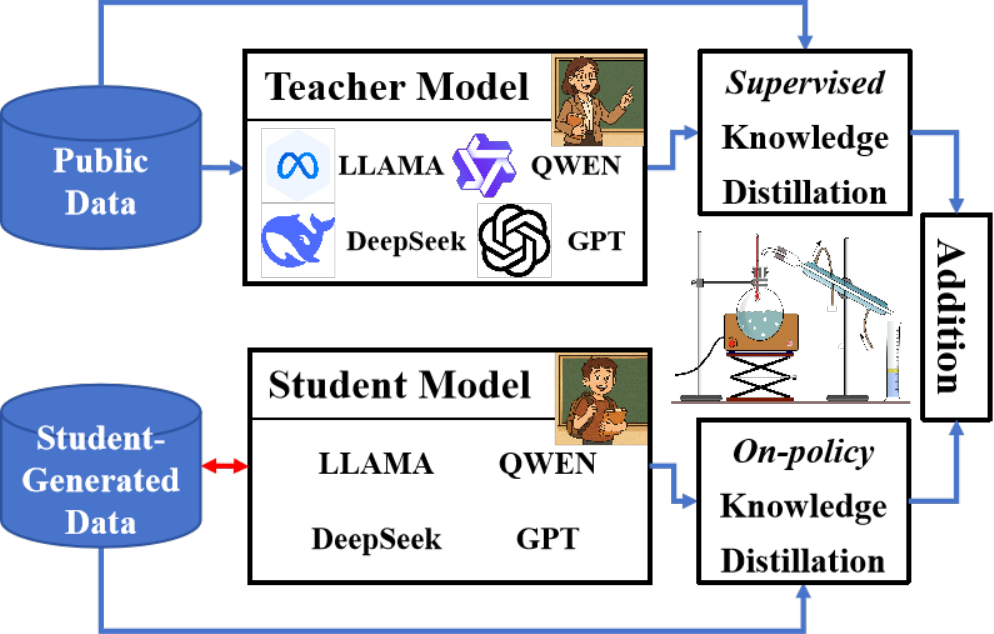}\label{fig:kd}}\hspace{0.1 cm}
	\subfigure[Compress-Then-Train]{\includegraphics[width=4.7 cm, height=4 cm]{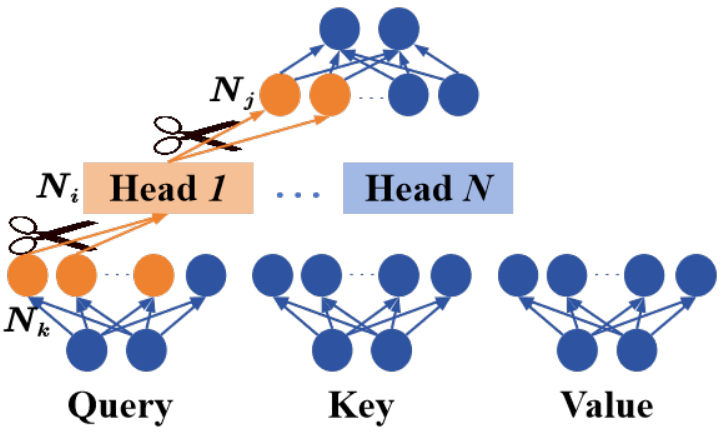}\label{fig:ctt}}\hspace{0.1 cm}
	\subfigure[One-Shot.]{\includegraphics[width=0.4\linewidth]{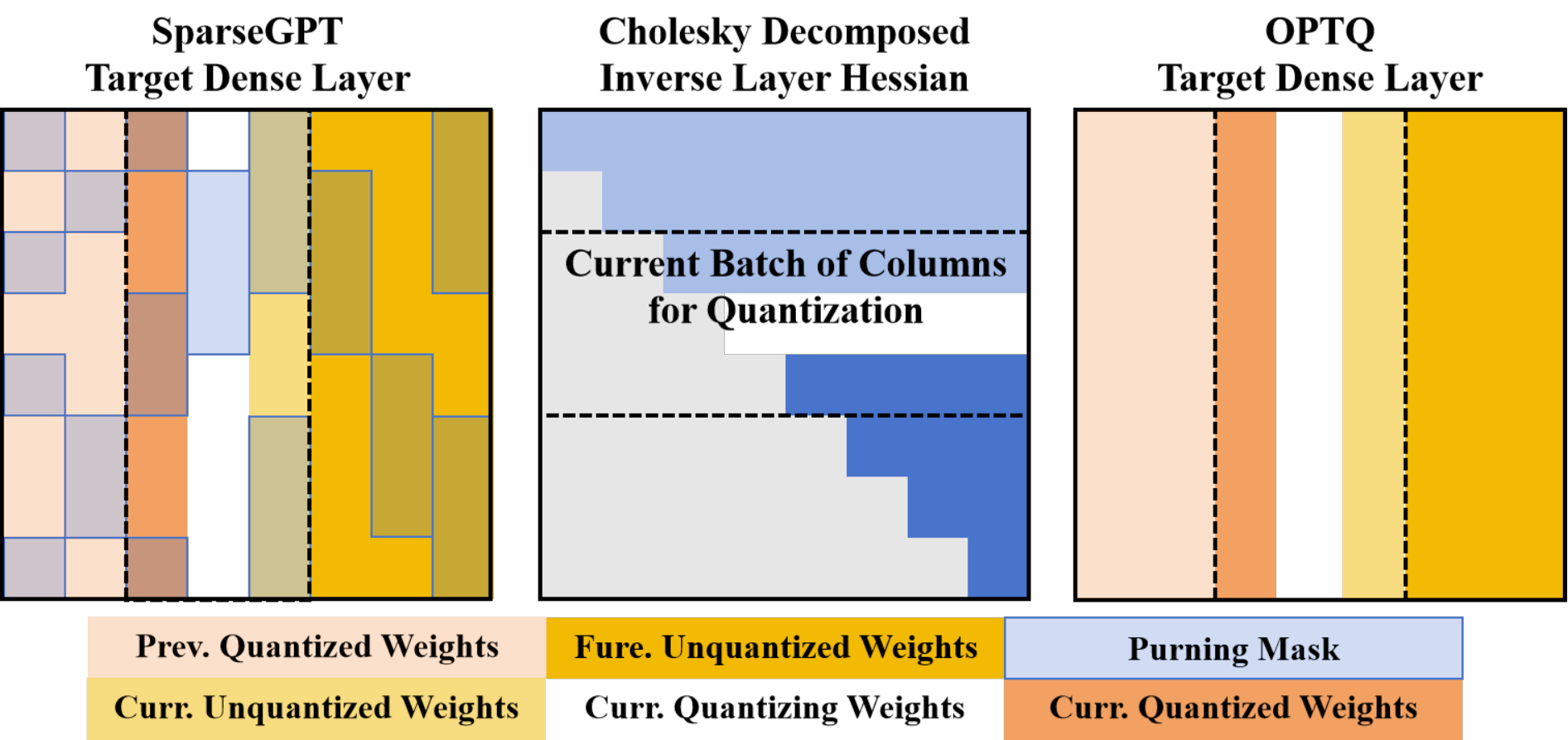}\label{fig:oneshot}}
	\vspace{-0.2 cm}
	\caption{\color{blue}The three categories of model compression techniques, i.e., compress-and-train, compress-then-train, and one-shot model compression.}\label{lbl:mc}
	\vspace{-0.2 cm}
	\hrulefill
	\vspace{-0.5 cm}
\end{figure*}

{\color{blue}
\paragraph{\bfseries Compression-and-Train}
In order to reduce the dependence on training data, a target LLM can be treated as a teacher model to distill and transfer the acquired knowledge to the student model that generally requires lower inference costs and memory footprint. 
However, traditional distillation techniques for LLMs require fixed datasets that necessitate either costly teacher-generated outputs or token-level probability annotations.
Due to limited expressiveness, student models may fail to match the distributions of teacher LLMs and generate unlikely samples when minimizing forward Kullback–Leibler (KL) divergence on the fixed datasets \cite{Agarwal2024}.
R. Agarwal \emph{et al.} in \cite{Agarwal2024} proposed a compress-and-train model compression, termed as generalized knowledge distillation (GKD), which handles the distribution mismatch based on: 1) replacing the forward KL divergence by the reverse KL divergence or generalized Jensen-Shannon (GJS) divergence; and 2) using the on-policy datasets that are generated by the student model under the guidance of teacher feedback, as shown in Fig. \ref{fig:kd}.
At the stage of distillation, the GKD technique combines the supervised distillation with the on-policy based distillation and can leverage the student-generated data (c.f. red arrow in Fig. \ref{fig:kd}) to mitigate the distribution mismatch between teacher and student models.
Moreover, the application of reverse KL and GJS divergence at distillation directs the student model toward the high-probability regions of teacher LLMs and prevents overfitting to low-probability regions in order to mitigate the limitations in expressiveness of student models.
Numerical results show that the GKD achieves substantial improvements in both factual consistency and summary quality when tested on the XSum dataset for summarization \cite{Agarwal2024}.
When distilled on task-agnostic fine-tuned language net dataset, the GKD with reverse KL divergence outperforms other distillation techniques on the massive multitask language understanding and big-bench hard benchmark suites, demonstrating its versatility and effectiveness for model performance preservation.

\paragraph{\bfseries Compression-Then-Train}
Despite its effectiveness, knowledge distillation typically requires a substantial amount of time to obtain a student model, e.g., 14 GPU days for the TinyBERT distillation \cite{Ma2023}.
To reduce retraining duration, model pruning is introduced to remove unnecessary neurons from LLMs while preserving the versatility.
In this vein, X. Ma \emph{et al.} proposed a task-agnostic model-pruning technique (termed as LLM-Pruner) that preserves multi-task versatility  without access to the original training data or lengthy retraining \cite{Ma2023}. 
The \mbox{LLM-Pruner} compresses the scale of LLMs with three stages, i.e., \emph{discovery}, \emph{estimation}, and \emph{recovery}. 
\textbf{Discovery stage} involves recursively activating all nodes that are dependent on the initial trigger node to obtain a group of coupled nodes. 
By applying such process to each node, the discovery stage can efficiently identifies all groups of coupled nodes as shown in Fig. \ref{fig:ctt}. 
Then, \textbf{estimation stage} evaluates the importance of each identified group based on the contribution to the  overall performance that can be estimated via the first-order and the approximated Hessian information.
Finally, \textbf{recovery stage} inserts trainable LoRA modules into all dense layers and employs first-order optimizers to rapidly restore model performance. 
Note that excessive tuning epochs can degrade performance \cite{Ma2023}. 
Each LoRA module operates in parallel with its corresponding dense layer, and after recovery, its weights are merged into the dense layer.
The three-stage LLM-Pruner can effectively reduce the model size and computational demands without significantly sacrificing performance. Even after pruning 20\% of the weights, the pruned LLMs retain nearly 95\% of the original performance with limited public data \cite{Ma2023}.

\paragraph{\bfseries One-Shot Compression}
Note that the retraining duration of the LLM-Pruner is still computationally expensive due to the iterative recovery stage. 
Therefore, the one-shot model compression techniques have been developed to reduce the scale of LLMs without retraining. 
Among the one-shot model compression techniques, one-shot quantization \cite{Frantar2023a} and the one-shot pruning \cite{Frantar2023} stand out due to their efficiency and effectiveness when applied to LLMs.

The open pre-trained transformer quantization (OPTQ) leverages the approximate Hessian information to achieve high accuracy and efficiency of one-shot compression \cite{Frantar2023a}.
The OPTQ initiates the quantization process by sequentially quantizing the weights of each layer in the LLM as shown in Fig. \ref{fig:oneshot}.
For each layer, the objective of OPTQ is to minimize mean-squared error (i.e., quantization error) between the full-precision output and the quantized one under {\color{blue}task-agnostic data} that can be crawled from the internet such that the performance loss of quantization can be mitigated. 
More specifically, OPTQ recursively selects a layer and simultaneously quantizes the selected columns of weight matrix.
Simultaneous quantization reduces the computational complexity associated with updating the inverse Hessian. 
However, due to numerical inaccuracies introduced by column-wise updates, the resulting inverse Hessian may become indefinite.
To ensure the positive-definiteness of the inverse Hessian, the OPTQ incorporates a Cholesky decomposition in the middle of Fig. \ref{fig:oneshot}.
By computing the inverse Hessian via the Cholesky decomposition and mild damping, the OPTQ can efficiently quantize weights without accumulating numerical errors.
To further enhance efficiency, the OPTQ employs a lazy batch-update strategy, i.e., updating the weights periodically instead of immediately per each quantization step. 
With the lazy batch-update strategy, the GPU utilization is improved by reducing the frequency of memory access.

The one-shot pruning (a.k.a., SparseGPT in \cite{Frantar2023}) recasts the pruning task as a series of layer-wise sparse regression via jointly selecting pruning masks and adjusting remaining weights as shown on the left side of  \ref{fig:oneshot}.  
Since simultaneous selection of pruning mask  and update of remaining weight is NP-hard, SparseGPT adopts a two-step approach: 1) adaptively pruning weights based on optimal brain surgeon error rank, and 2) minimizing the layer-wise mean-squared error between the outputs of the original and pruned layers.
During the pruning mask selection, the $p\%$ of weights with the smallest optimal-brain-surgeon errors are marked for pruning.
By incorporating the pruning mask, the SparseGPT applies weight-quantization procedures similar to those used in OPTQ.
Moreover, the sparsification and weight quantization can be integrated with negligible additional computational cost to efficiently prune the target LLMs.

The above model-compression techniques provide a range of options depending on the desired trade-off between efficiency, accuracy, and computational overhead.
Compression-and-train techniques (e.g., GKD \cite{Agarwal2024}) focus on transferring knowledge from teacher to student models and excel in versatility but require significant training resources. 
Compression-then-train approaches (e.g., LLM-Pruner \cite{Ma2023}) reduce retraining time and maintain multi-task performance through structured pruning and efficient recovery though recovery still incurs computational costs. 
One-shot compression techniques (e.g., OPTQ \cite{Frantar2023a} and SparseGPT \cite{Frantar2023}) perform quantization or pruning without retraining and can offer rapid and resource-efficient deployment.
However, the one-shot compression techniques may suffer from lower peak performance than techniques with post-compression training.
}

\section{Conclusions and Future Research Directions}
Given the substantial computational and memory demands of LLM inference and deployment, federated PEFT, federated MEFT, and model-compression techniques are indispensable for bringing LLM capabilities to resource-constrained edge clients.
Our work provided a comprehensive overview of the recent advances on federated PEFT, federated MEFT, and model compression to highlight their main contributions. 
We conclude our work with three promising research directions that warrant further exploration.

\paragraph{\bfseries Privacy-Preserving}While both federated PEFT and federated MEFT techniques in DLNoE avoid direct exchange of training data, LLMs are still susceptible to memorizing user-specific and domain-specific information in the fine-tuning datasets. 
Such unintended memorization poses a significant risk of exposing sensitive content during model inference. 
Therefore,  privacy-preserving mechanisms must be systematically engineered to mitigate memorization-induced leakage, thereby reinforcing the confidentiality of user data.

{\color{blue}
\paragraph{\bfseries Model Security}The federated MEFT techniques exchange only the SG dictionary during fine-tuning, which increases their vulnerability to eavesdropping and Byzantine attacks.
Intercepted SG dictionaries can enable reconstruction of proprietary LLMs and threaten data confidentiality. 
Moreover, the compromised seeds hinder clients from reproducing perturbations and thereby disrupt collaboration. 
Therefore, future research should comprehensively investigate defense mechanisms to safeguard federated MEFT against both eavesdroppers and Byzantine adversaries.
}

\paragraph{\bfseries Module Initialization}The performance gap between full-parameter fine-tuning and PEFT can be large due to the randomly initialized LoRA weights.
For example, LoRA-based FedKSeed underperforms the vanilla FedKSeed. 
An important future research direction for federated MEFT in DLNoE scenarios, therefore, is the development of principled initialization schemes for LoRA modules to accelerate LoRA adaptation and narrow the gap to full fine-tuning.

%\vspace{-0.1 cm}
\bibliographystyle{IEEEtran}
\bibliography{new_RL}

\end{document}